\documentclass[twocolumn,times,final,authoryear]{elsarticle}
\usepackage[latin9]{inputenc}
\usepackage{float}
\usepackage{amsmath}
\usepackage{amssymb}
\usepackage{graphicx}

\makeatletter

\providecommand{\tabularnewline}{\\}
\newcommand{\lyxdot}{.}

\floatstyle{ruled}
\newfloat{algorithm}{tbp}{loa}
\providecommand{\algorithmname}{Algorithm}
\floatname{algorithm}{\protect\algorithmname}

\usepackage{ycviu2}
\usepackage{framed}
\usepackage{multirow}

\usepackage{latexsym}

\usepackage{url}
\usepackage{xcolor}
\definecolor{newcolor}{rgb}{.8,.349,.1}

\makeatother

\begin{document}
\clearpage{}

\ifpreprint \setcounter{page}{1} \else \setcounter{page}{1} \fi

\begin{frontmatter}

\title{Neural Generative Models for 3D Faces with Application in 3D Texture
Free Face Recognition}

\author[1]{Ahmed ElSayed\corref{cor1}}

\cortext[cor1]{Corresponding author: Tel.: +0-000-000-0000; fax: +0-000-000-0000;}

\ead{aelsayed@my.bridgeport.edu}

\author[2]{Elif Kongar}

\author[1]{Ausif Mahmood}

\author[1]{Tarek Sobh}

\author[3]{Terrance Boult}

\address[1]{Department of Computer Science and Engineering, School of Engineering,
University of Bridgeport, Bridgeport, CT 06604, USA}

\address[2]{Department of Mechanical Engineering and Technology Management,
School of Engineering, University of Bridgeport, Bridgeport, CT 06604,
USA}

\address[3]{El Pomar Institute for Innovation and Commercialization (EPIIC),
University of Colorado, Colorado Springs, CO 80918, USA}
\begin{abstract}
Using heterogeneous depth cameras and 3D scanners in 3D face verification
causes variations in the resolution of the 3D point clouds. To solve
this issue, previous studies use 3D registration techniques. Out of
these proposed techniques, detecting points of correspondence is proven
to be an efficient method given that the data belongs to the same
individual. However, if the data belongs to different persons, the
registration algorithms can convert the 3D point cloud of one person
to another, destroying the distinguishing features between the two
point clouds. Another issue regarding the storage size of the point
clouds. That is, if the captured depth image contains around 50 thousand
points in the cloud for a single pose for one individual, then the
storage size of the entire dataset will be in order of giga if not
tera bytes. With these motivations, this work introduces a new technique
for 3D point clouds generation using a neural modeling system to handle
the differences caused by heterogeneous depth cameras, and to generate
a new face canonical compact representation. The proposed system reduces
the stored 3D dataset size, and if required, provides an accurate
dataset regeneration. Furthermore, the system generates neural models
for all gallery point clouds and stores these models to represent
the faces in the recognition or verification processes. For the probe
cloud to be verified, a new model is generated specifically for that
particular cloud and is matched against pre-stored gallery model presentations
to identify the query cloud. This work also introduces the utilization
of Siamese deep neural network in 3D face verification using generated
model representations as raw data for the deep network, and shows
that the accuracy of the trained network is comparable all published
results on Bosphorus dataset.
\end{abstract}
\end{frontmatter}



\section{Introduction}

Rapid improvements in 3D capturing techniques increased the utilization
of 3D face recognition especially when the regular 2D images fail
due to lighting and appearance changes. The techniques used for 3D
based face recognition have been summarized in \citep{3D-survey,3D-survey2,3D-faces-book}.
Relevant studies are explained in the following text.

The work in \citep{7-3D} uses 3D face recognition by segmenting a
range image based on principal curvature and finding a plane of bilateral
symmetry through the face. This plane is used for pose normalization.
The authors consider methods of matching the profile from the plane
of symmetry and of matching the face surface. A modified technique
proposed in \citep{12-3D}, where the authors use segment convex regions
in the range image based on the sign of the mean and Gaussian curvatures,
and create an Extended Gaussian Image (EGI) for each convex region.
A match between a region in a probe image and in a gallery image is
done by correlating EGIs. A graph matching algorithm incorporating
relational constraints is used to establish an overall match of probe
image to gallery image. Convex regions are believed to change shape
less than other regions in response to changes in facial expression.
This gives this approach some ability to cope with changes in facial
expression. However, EGIs are not sensitive to change in object size,
and hence two similarly shaped differently sized faces will not be
distinguishable in this representation. In \citep{9-3D} the author
begins with a curvature-based segmentation of the face. Then a set
of features are extracted that describe both curvature and metric
size properties of the face. Thus each face becomes a point in feature
space, and matching is done by a nearest-neighbor match in feature
space. It is noted that the values of the features used are generally
similar for different images of the same face, \textquotedblleft except
for the cases with large feature detection error, or variation due
to expression\textquotedblright{} \citep{9-3D}. Instead of working
on all face points, in \citep{18-3D} the authors used 3D five feature
points only, using these feature points to standardize face pose,
and then matching various curves or profiles through the face data.
Experiments are performed for sixteen subjects, with ten images per
subject. The best recognition rates are found using vertical profile
curves that pass through the central portion of the face. Computational
requirements were apparently regarded as severe at the time this work
was performed, as the authors note that \textquotedblleft using the
whole facial data may not be feasible considering the large computation
and hardware capacity needed\textquotedblright{} \citep{18-3D}. In
\citep{3-3D} they extend Eigenface and hidden Markov model approaches
used for 2D face recognition to work with range images. \citep{20-3D}
also perform curvature-based segmentation and represent the face using
an Extended Gaussian Image (EGI). Recognition is then performed using
a spherical correlation of the EGIs. In \citep{2-3D} the authors
report on a method of 3D face recognition that uses an extension of
the Hausdorff distance matching. Again, work in \citep{10-3D} explores
principal component analysis (PCA) style approaches using different
numbers of eigenvectors and image sizes. The image data set used has
6 different facial expressions for each of the 37 subjects. The performance
figures reported resulting from using multiple images per subject
in the gallery. This effectively gives the probe image more chances
to make a correct match, and is known to raise the recognition rate
relative to having a single sample per subject in the gallery \citep{16-3D}.
Registration and correspondence has been used in \citep{14-3D} to
perform 3D face recognition using iterative closest point (ICP) matching
of face surfaces. Even though most of the work covered here used 3D
shape acquired through a structured-light sensor, this work uses a
stereo-based system. The Approach used in \citep{17-3D} is 3D face
recognition by first performing a segmentation based on Gaussian curvature
and then creating a feature vector based on the segmented regions.
The authors report results on a dataset of 420 face meshes representing
60 different persons, with some sampling of different expressions
and poses for each person. Another research is perform 3D face recognition
by locating the nose tip, and then forming a feature vector based
on contours along the face at a sequence of depth values \citep{13-3D}.
An isometric transformation approach has been used in \citep{6-3D}
to analyze 3D face in an attempt to better cope with variation due
to facial expression. Rather than performing recognition on the all
face as one module, the authors in \citep{Bosph_results} have performed
recognition using registration on separate face parts and uses fusion
to come up with a final decision. Moreover, other research as in \citep{Bosph_best}
and \citep{mesh-SIFT} use the high dimensional extracted features,
viz. scale invariant feature transform (SIFT, mesh-SIFT) and histograms
for both gradient and shapes, from the 3D cloud, and perform the recognition
process on them. 

\begin{figure}
\begin{centering}
\includegraphics[scale=0.3]{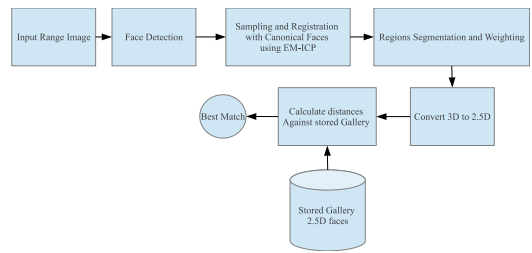}
\par\end{centering}
\caption{Example for 3D recognition using registration and projection to 2.5D.
\label{fig:Example-for-3D}}
\end{figure}

The survey indicates that most if not all research work on extracting
some features from given 3D face point clouds and using these features
in the recognition process. The extracted features heavily depend
on the cloud space and can be easily affected by the structure and
size of the given points cloud. These clouds can be modeled to save
the storage size or to regenerate the depth information. Other research
converts the given 3D points cloud to 2.5D at standard $X$ and $Y$
coordinates using orthographic projection and convert the problem
to pixel image recognition as suggested by \citep{realtime-3D-face}.
Figure \ref{fig:Example-for-3D} shows a complete system that uses
this technique. 

\begin{figure}
\begin{centering}
\includegraphics[scale=0.3]{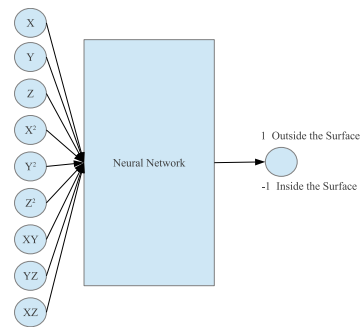}
\par\end{centering}
\caption{Neural network for 3D points modeling in Cretu et al. \label{fig:Neural-network-for}}
\end{figure}

\begin{figure}
\begin{centering}
\includegraphics[scale=0.4]{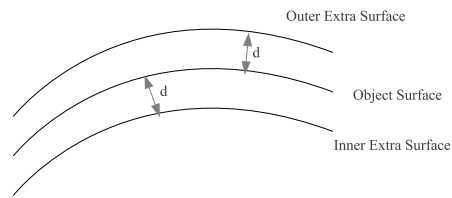}
\par\end{centering}
\caption{Extra surfaces generated for neural model learning in Cretu et al.
\label{fig:extra-surfaces}}
\end{figure}

As previously mentioned, there is literature that focus on modeling
the 3D face clouds as in \citep{neural-model}. Here, a neural model
is designed as shown in Figure \ref{fig:Neural-network-for}. In this
network, the input consists of second order values for all input points
cloud and the output is 0. Additional input values are added for extra
generated surfaces inside and outside the point cloud surface. The
output in this case should be proportional to the distance from the
input point to the cloud surface as shown in Figure \ref{fig:extra-surfaces}.
The model however, requires the generation of at least 5 times the
number of the original points cloud. Furthermore, the output is not
guaranteed to be on the original surface since the acceptance tolerance
$d$ is defined for scaling purposes making this model computationally
expensive if a higher accuracy required. 

Despite the fact that there is an increase in the literature that
includes Deep-learning and Deep-neural systems in 3D object recognition
as in \citep{3D-CRNN,3D-object-CNNraey,deep-belief-3D}, none of these
techniques have been applied to 3D face recognition, or it can use
the regular 2D representation of the 3D face as in \citep{Deep3DID}.
One reason for this lack of applicability is the high sensitivity
and the closeness of features between the faces of different individuals,
specifically if the 3D data is used alone without any texture. 

Addressing these issues, the main contributions of the proposed 3D
neural recognition and verification system are listed as follows. 
\begin{enumerate}
\item Designing neural generative model for representation and reconstruction
of 3D faces,
\item Significantly reducing the storage space used for the 3D point clouds,
by replacing the stored point clouds with the generated neural model
representations,
\item Using the generated presentation from the 3D regression models of
gallery set for recognition and verification against generated model
representation for probe points cloud,
\item Combining generated face model representation with Siamese network
to generate a comprehensive framework for 3D face verification. 
\end{enumerate}

\section{Proposed 3D Based Face Recognition System}

\begin{figure}
\begin{centering}
\includegraphics[scale=0.3]{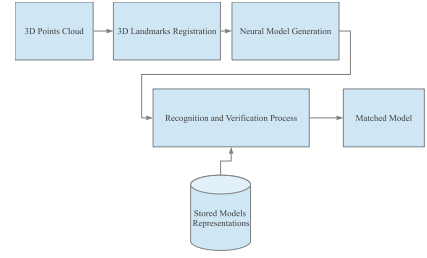}
\par\end{centering}
\caption{Proposed 3D based face recognition system. \label{fig:Proposed-Depth-based}}
\end{figure}

In some cases, due to lighting conditions and/or makeup or other 2D
effects in the face image, the regular 2D image becomes insufficient
for face recognition. In order to address these issues, this work
presents a 3D based face recognition system that is able to work on
the texture free 3D point clouds extracted by depth cameras to identify
or verify the person. In this regard, this research introduces a new
technique for 3D cloud regeneration using a neural generative model
to handle the differences caused by heterogeneous depth cameras, and
to generate a new face canonical compact representation. The proposed
system reduces the stored 3D dataset size and if required, provides
an accurate dataset regeneration. Furthermore, the system generates
neural models for all gallery point clouds and stores these models
to represent these faces in the recognition or verification process.
For the probe cloud to be verified, the system obtains the 3D points
cloud as an input with face landmark points. These landmark points
are then registered to reference points to align and to scale the
input cloud correctly. After the registration step, a neural model
is generated for this prob cloud to provide a compact representation
of the 3D face data. The extracted neural model is then applied to
a face recognition or verification step to detect the best matched
model from the pre-stored gallery model presentations. This work also
introduces the utilization of Siamese deep neural network in 3D face
verification using generated model representation as a raw data for
the deep network. The complete proposed system is depicted in Figure
\ref{fig:Proposed-Depth-based}. The following sections will explain
each step used in the system.

\section{3D Registration }

The first step in the proposed system is the registration that transforms
all 3D data into a canonical standard position. This step involves
3D data using Iterative Closest Point (ICP) algorithm. Before applying
ICP on the input clouds, these clouds have to be normalized around
their means and to be placed inside a cube with maximum dimensions
of $2\times2\times2$ with each axis having a range $\left[-1,1\right]$.
After the normalization step, the landmark points registration is
applied to obtain the suitable rigid transformation between input
landmark points and the reference points. The output transformation
is then applied to all input points to obtain the registered points
which will be used in the following steps. 

\subsection{Iterative Closest Point (ICP)}

Iterative Closest Point (ICP), a.k.a., Iterative Corresponding Point,
is an algorithm used to obtain the corresponding points and transformations
between two groups of points in 2D or 3D. However, the algorithm is
mostly used in the 3D cases for registration between some query mesh
and standard canonical mesh. The main algorithm was introduced in
\citep{ICP}. The main goal of \citep{ICP} work is to obtain the
optimum transformation matrix $T=\left[R|t\right]$ (where $R$ is
the $3\times3$ rotation matrix and $t$ is $3\times1$ translation
vector) that can convert moving landmark points set $M=\left\{ m_{i}\right\} $
to the static points set $S=\left\{ s_{i}\right\} $. Assuming that
the number of points for both sets are equal ($N_{M}=N_{s}$), the
objective function required to be minimized should be

\begin{equation}
f\left(T\right)=\frac{1}{N_{S}}\sum_{i=1}^{N_{s}}\left\Vert s_{i}-Rm_{i}-t\right\Vert ^{2}.\label{eq:minimzed-equation}
\end{equation}

To achieve this goal the following steps are followed:
\begin{enumerate}
\item Starting from iteration $k=0$, $M_{k}=M_{0}=M$
\item Find the closest corresponding points (on Euclidean space) between
$M_{k}$ set and $S$ set. These correspondence set will be $Y_{k}=C\left(M_{k},S\right)$.
\item For the particular $Y_{k}$, minimize equation \eqref{eq:minimzed-equation}
solving for the value of $T$ (using least square techniques). The
solution will be $T_{k}$
\item Apply $T_{k}$over $M_{k}$ set to generate $M_{k+1}=R_{k}M_{k}+t_{k}$
\item Set $k=k+1$
\item Repeat steps 2,3 and 4 until the stopping criteria are satisfied.
\end{enumerate}
The main disadvantage of this technique is its computational complexity
which reaches to significant levels when the number of points are
large ($O\left(N_{m}N_{s}\right)$). However, some research uses other
techniques rather than Nearest Neighborhood (NN), which used in step
2, to improve the computation. K-D Tree is one of these alternative
algorithms that can be used for this improvement. Other modified versions
of ICP have also been introduced to make the algorithm more computationally
efficient. As stated in \citep{fast-ICP} these improvements have
been summarized in the following steps:
\begin{enumerate}
\item Selecting some set of points in one or both sets. 
\item Matching these points to samples in the other set. 
\item Weighting the corresponding pairs appropriately. 
\item Rejecting certain pairs by looking at each pair individually or considering
the entire set of pairs. 
\item Assigning an error metric based on the point pairs. 
\item Minimizing the error metric.
\end{enumerate}
Based on the new steps, the algorithm will not work on these entire
sets, but on some selected samples from both sets. The sampling and
rejection steps in the modified algorithm improve the computation
complexity, even though this new approach leads to other concerns
regarding the best sampling and rejection mechanisms that can be used
to obtain the best matching result.

\section{Neural Regression Model for 3D Face Representation}

\begin{figure}
\begin{centering}
\includegraphics[scale=0.3]{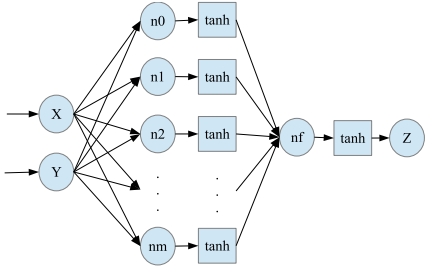}
\par\end{centering}
\caption{Proposed neural representation of face depth data .\label{fig:Proposed-Neural-Representation}}
\end{figure}

As mentioned previously, one of the problems in the 3D point clouds
is the storage size. The storage size for one 3D face including about
80,000 points is in the order of tens if not hundreds of mega bytes.
Therefore, finding an alternative representation that can reduce the
size while providing the same accuracy is important. An additional
concern is th different 3D cameras will produce different cloud resolution
(different 3D sensors produce different number of 3D points for the
same object). This concern is valid regardless of that the image representing
a single person or multiple individuals. To address these issues,
a new neural representation is proposed as shown in Figure \ref{fig:Proposed-Neural-Representation}.
The proposed neural model will obtain $X$ and $Y$ coordinates as
input from the points cloud and will generate the corresponding $\bar{Z}$
values for the same points, which should represent the actual $Z$
values of these points. The mathematical representation of the model
will be as the following equations:

\begin{equation}
\bar{Z}^{i}=\tanh\left(n_{f}^{i}\right),
\end{equation}

\begin{equation}
n_{f}^{i}=\sum_{j=1}^{M}Wo_{jf}\cdot\tanh\left(n_{j}^{i}\right)+Bo_{f},
\end{equation}
\begin{equation}
n_{j}^{i}=\left(Wi_{1j}\cdot X^{i}+Wi_{2j}\cdot Y^{i}\right)+Bi_{j},
\end{equation}

where $M$ is the number of hidden units, $\left(X^{i},Y^{i},Z^{i}\right)$
is a 3D point in the point cloud and its corresponding neural output
$\bar{Z}^{i}$ and $j$ is the hidden node index. 

To obtain the weights $Wi$ and $Wo$ for the neural model, a loss
function should be defined and optimized. In this work, the Mean Squared
Error (MSE) will be used as stated in equation \eqref{eq:depth-model-loss}.

\begin{equation}
\mathcal{L}\left(P\right)=\frac{1}{N}\sum_{i=1}^{N}\left\Vert \bar{Z}^{i}\left(X^{i},Y^{i},P\right)-Z^{i}\right\Vert {}^{2}.\label{eq:depth-model-loss}
\end{equation}

where $P=\left\{ Wi_{1j},Wi_{2j},Bi_{j},Wo_{jf},Bo_{f}\right\} $,
$N$ is the number of samples in the points cloud, $Wi_{1,j}$ is
the weight of the hidden node $j$ for the first input (the $X$ coordination),
$Wi_{2,j}$ is the weight of the hidden node $j$ for the second input
(the $Y$ coordination), $Bi_{j}$ is the bias of the hidden node
$j$ and $Bo_{f}$ is the bias of the final output node.

To solve this optimization problem, Levenberg\textendash Marquardt
Back-Propagation (LM) will be used to obtain the values of $P$ that
provides the minimum Mean Squared Error (MSE).

The main advantages of this proposed model are:
\begin{itemize}
\item Can be used as a raw data for recognition or verification process:
as will be explained and tested in the next sections, the weight matrix
$P$ can be flattened (converted to 1D vector) and used as a raw data
for verification and recognition.
\item Easer in data augmentation (if the model is used as a raw data features
for the verification process): in various machine learning problems,
the number of available samples for training or testing are limited.
To overcome this issue, data augmentation is commonly used to generate
additional samples by manipulating the existing data (via shifting,
rotating and clipping in the case of image learning). However, in
3D cases, especially for faces, these processes can generate limited
number of samples. For the proposed model however, significantly large
numbers of new models can be generated for the same face by only swapping
rows in the matrix $P$. For example, assuming the structure of the
network is 2-500-1 (2 is the number of inputs ($X$ and $Y$ coordinates),
$M$ (number of hidden layers) is 500 and 1 output ($Z$ coordinate)),
then the size of $Bo_{f}$ will be $1\times1$ which can be concatenated
to the matrix $P$ after flatting it, and the value of $B_{f}$ will
not change in all augmented models generated from the original model
$P$. So if the original trained model weights $P=P_{1}=\left[\begin{array}{cccc}
wi_{1,1} & wi_{2,1} & bi_{1} & wo_{1,1}\\
wi_{1,2} & wi_{2,2} & bi_{2} & wo_{2,1}\\
wi_{1,3} & wi_{2,3} & bi_{3} & wo_{3,1}\\
\vdots & \vdots & \vdots & \vdots\\
wi_{1,500} & wi_{2,500} & bi_{500} & wo_{500,1}
\end{array}\right]$ then another model $P_{2}=\left[\begin{array}{cccc}
wi_{1,3} & wi_{2,3} & bi_{3} & wo_{3,1}\\
wi_{1,2} & wi_{2,2} & bi_{2} & wo_{2,1}\\
wi_{1,1} & wi_{2,1} & bi_{1} & wo_{1,1}\\
\vdots & \vdots & \vdots & \vdots\\
wi_{1,500} & wi_{2,500} & bi_{500} & wo_{500,1}
\end{array}\right]$ can be generated as a new model that represents the same face as
$P_{1}$ by only swapping rows 1 and 3. Therefore, assuming that $M$
equal to 500, this technique can generate $500!$ different models
for the same face from a single trained model. 
\item Reduce the storage size of the face representation: because to store
a 3D face only the network weights need to be stored (assuming the
network structure is known). For instance, if the network structure
is (2-500-1), the number of stored weights will be 2001 (taking into
account the network biases). This means that instead of storing about
80,000 double precision numbers or more (the original number of points
in the cloud), only 2,000 float precision numbers can be stored which
will improve the storage size with a factor of 80.
\item Independent of the number of points in the trained cloud: to train
the proposed neural model you don't need specific number of points
in the cloud (you can train the model with 80,000, 50,000 or 15,000
points). Which means that the generated model can be trained using
different camera's resolution. This makes this technique suitable
for heterogeneous cameras.
\item Can be used in 3D super-resolution: the neural generated model designed
in this work is a regression model that can be used for smoother accurate
interpolation to generate higher resolution version from the original
3D points, which can considered as 3D super-resolution algorithm. 
\end{itemize}

\section{Levenberg\textendash Marquardt Back-Propagation}

In the Gradient Decent Back-Propagation algorithm, the back-propagated
recurrent form of sensitivity at layer $k$ in the neural network
has been formulated as 

\begin{equation}
\delta^{k}=\dot{F}^{k}\left(n^{k}\right).W^{k+1^{T}}.\delta^{k+1},\label{eq:recurrent-sens}
\end{equation}

where

\begin{equation}
\dot{F}^{k}\left(n^{k}\right)=\left[\begin{array}{cccc}
\dot{f}^{k}\left(n^{k}\left(1\right)\right) & 0 & \cdots & 0\\
0 & \dot{f}^{k}\left(n^{k}\left(2\right)\right) & \cdots & 0\\
\vdots & \vdots & \ddots & \vdots\\
0 & 0 & \cdots & \dot{f}^{k}\left(n^{k}\left(S_{k}\right)\right)
\end{array}\right],\label{eq:complete-drivative}
\end{equation}

and

\begin{equation}
\dot{f}^{k}\left(n^{k}\left(i\right)\right)=\frac{df^{k}\left(n^{k}\left(i\right)\right)}{dn^{k}\left(i\right)}.\label{eq:drivative}
\end{equation}

where $f^{k}\left(n^{k}\left(i\right)\right)$ is the activation function
of node $i$ in layer $k$, $n^{k}\left(i\right)$ is the output of
node $i$ in layer $k$ and $W$ represents all weights and biases
in the network.

The Gradient Decent method works well when the network task is a classification
with softmax loss function. However, for some tasks, the loss function
is defined as Mean Squared Error (MSE) function between neural network
output $a^{M}$and required output $y$. Based on this, the loss function
can be defined as

\begin{equation}
e_{i}\left(W\right)=\left(y_{i}-a_{i}^{M}\left(W\right)\right),\label{eq:error}
\end{equation}

\begin{equation}
\mathcal{L}\left(W\right)=\frac{1}{2}\sum_{i=1}^{N}e_{i}^{2}\left(W\right),
\end{equation}

where $N=Q\times S_{m}$

Based on Levenberg\textendash Marquardt algorithm used in \citep{LM-Back},
weights and biases update can be calculated as

\begin{equation}
\Delta W=-\left[\nabla^{2}\mathcal{L}\left(W\right)\right]^{-1}\nabla\mathcal{L}\left(W\right),\label{eq:update}
\end{equation}

where $\nabla^{2}\mathcal{L}\left(W\right)$ is the Hessian matrix
and $\nabla\mathcal{L}\left(W\right)$is the gradient. The gradient
term can be expressed as

\begin{equation}
\nabla\mathcal{L}\left(W\right)=J^{T}\left(W\right)e\left(W\right),\label{eq:grad}
\end{equation}

where $e\left(w\right)=\left[\begin{array}{c}
e_{1}\left(W\right)\\
e_{2}\left(W\right)\\
\vdots\\
e_{N}\left(W\right)
\end{array}\right]$, and the Hessian matrix can be approximated as 

\begin{equation}
\nabla^{2}\mathcal{L}\left(W\right)=J^{T}\left(W\right)J\left(W\right),\label{eq:hessian}
\end{equation}

where $J$ is the Jacobian matrix stated as 
\begin{equation}
J\left(W\right)=\left[\begin{array}{cccc}
\frac{\partial e_{1}\left(W\right)}{\partial W_{1}} & \frac{\partial e_{1}\left(W\right)}{\partial W_{2}} & \cdots & \frac{\partial e_{1}\left(W\right)}{\partial W_{n}}\\
\frac{\partial e_{2}\left(W\right)}{\partial W_{1}} & \frac{\partial e_{2}\left(W\right)}{\partial W_{2}} & \cdots & \frac{\partial e_{2}\left(W\right)}{\partial W_{n}}\\
\vdots & \vdots & \ddots & \vdots\\
\frac{\partial e_{N}\left(W\right)}{\partial W_{1}} & \frac{\partial e_{N}\left(W\right)}{\partial W_{2}} & \cdots & \frac{\partial e_{N}\left(W\right)}{\partial W_{n}}
\end{array}\right],\label{eq:jacobian}
\end{equation}

From equations \eqref{eq:grad}, \eqref{eq:hessian}, \eqref{eq:jacobian}
and \eqref{eq:update} can be expressed as 

\begin{equation}
\Delta W=\left[J^{T}\left(W\right)J\left(W\right)\right]^{-1}J^{T}\left(W\right)e\left(W\right),
\end{equation}

Adding one more control parameter the update can be stated as

\begin{equation}
\Delta W=\left[J^{T}\left(W\right)J\left(W\right)+\mu I\right]^{-1}J^{T}\left(W\right)e\left(W\right),\label{eq:LM_update}
\end{equation}

The new added parameter $\mu$ is used as a variable step control
for the updates based on the loss value. Every time the loss $\mathcal{L}\left(W\right)$
reduces, the value of $\mu$ is divided over some other constant parameter
$\beta$ to go closer to the minimum loss value.

Applying the new equation to back-propagation algorithm in section
3 resulting in new weight update equation 

\begin{equation}
\Delta w^{k+1}\left(i,j\right)=-\alpha.\frac{\partial L_{q}}{\partial w^{k+1}\left(i,j\right)}=-\alpha.\frac{\partial\sum_{m=1}^{S_{M}}e_{q}^{2}\left(m\right)}{\partial w^{k+1}\left(i,j\right)},
\end{equation}

\begin{equation}
\Delta b^{k+1}\left(i\right)=-\alpha.\frac{\partial L_{q}}{\partial b^{k+1}\left(i\right)}=\Delta b^{k+1}\left(i\right)=-\alpha.\frac{\partial\sum_{m=1}^{S_{M}}e_{q}^{2}\left(m\right)}{\partial b^{k+1}\left(i\right)},
\end{equation}

Identical steps used in the regular back-propagation can also be used
with the LM method to fill the Jacobian matrix with small modification
at the final step
\begin{equation}
\delta^{M}=-\dot{F}^{M}\left(n^{M}\right).\label{eq:LM-final-sens}
\end{equation}

Based on these equations, the LM back-propagation algorithm will work
as follows:

\begin{algorithm}
Apply all $Q$ inputs to the network and calculate network outputs
and errors corresponding to these output and loss value.
\begin{enumerate}
\item Use equations \eqref{eq:LM-final-sens}, \eqref{eq:drivative}, \eqref{eq:complete-drivative},
\eqref{eq:recurrent-sens}, \eqref{eq:jacobian} to calculate Jacobian
matrix (other efficient Jacobian calculation methods can be used in
this step).
\item Solve equation \eqref{eq:LM_update} (using Cholesky factorization)
for the $\Delta W$.
\item Use the calculated $\Delta W$ to calculate the new value of $W+\Delta W$.
\item Check the new loss value, if the loss value decrease, then decrease
$\mu$ by $\beta$ , update $W=W+\Delta W$ and go to step 1. If the
loss didn't reduce increase $\mu$ by $\beta$ and go to step 3.
\item Repeat the these steps until the stopping criteria are satisfied.
\end{enumerate}
\caption{Levenberg\textendash Marquardt algorithm}
\end{algorithm}

\section{Recognition and Verification }

Using similarity metric for face verification is proven to be an efficient
method, especially if the number of images per class is low. Therefore,
this work utilizes the similarity metric method with Convolutional
Neural Network (CNN) to perform a Siamese Network that will be applied
in the final step of the proposed system to perform the verification
process.

\begin{figure}
\begin{centering}
\includegraphics[scale=0.3]{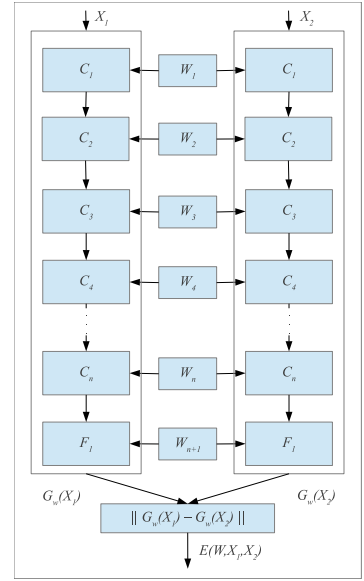}
\par\end{centering}
\caption{Typical structure of Siamese network. \label{fig:Typical-Siamese}}
\end{figure}

Using Siamese Network for face verification has been introduced in
\citep{siamese}, where two input images $X_{1}$ and $X_{2}$ are
applied to the same nonlinear mapping $G_{W}$ to extract the main
features that minimize the main energy function $E$ when $X_{1}$
and $X_{2}$ belong to the same person and maximize it when they belong
to different persons. The typical structure for this network is shown
in Figure \ref{fig:Typical-Siamese} \citep{siamese}. The formal
definition of the function $E$ can be expressed as in equation \eqref{eq:Siamese_energy}

\begin{equation}
E\left(W,X_{1},X_{2}\right)=\left\Vert G_{W}\left(X_{1}\right)-G_{W}\left(X_{2}\right)\right\Vert ,\label{eq:Siamese_energy}
\end{equation}

where $W$ are the shared weight filters between the two input images.

To achieve this goal for the $E$ function, the loss function should
monotonically increase with same person pairs' energy and monotonically
decrease with different persons pairs' energy. Based on this, the
final loss function will be formed as in equations \eqref{eq:Siamese_loss},\eqref{eq:Siamese_iter_loss},\eqref{eq:Same_loss},\eqref{eq:diff_loss},\eqref{eq:siamese_out}

\begin{equation}
\mathcal{L}\left(W\right)=\sum_{i=1}^{N}L\left(W,\left(Y,X_{1},X_{2}\right)^{i}\right),\label{eq:Siamese_loss}
\end{equation}

\begin{multline}
L\left(W,\left(Y,X_{1},X_{2}\right)^{i}\right)=Y\cdot L^{s}\left(E\left(W,X_{1},X_{2}\right)^{i}\right)+\\
\left(1-Y\right)\cdot L^{d}\left(E\left(W,X_{1},X_{2}\right)^{i}\right),\label{eq:Siamese_iter_loss}
\end{multline}

\begin{equation}
L^{s}\left(E\left(W,X_{1},X_{2}\right)^{i}\right)=\frac{2}{Q}\left(E\left(W,X_{1},X_{2}\right)^{i}\right)^{2},\label{eq:Same_loss}
\end{equation}

\begin{equation}
L^{d}\left(E\left(W,X_{1},X_{2}\right)^{i}\right)=2Q\cdot e^{\left(-\frac{2.77}{Q}E\left(W,X_{1},X_{2}\right)^{i}\right)},\label{eq:diff_loss}
\end{equation}

\begin{equation}
Y=\begin{cases}
1 & X_{1}\equiv X_{2}\\
0 & X_{1}\not\equiv X_{2}
\end{cases}.\label{eq:siamese_out}
\end{equation}

where $N$ is the number of training samples, $Y$ is equal to $1$
if $X_{1}$ and $X_{2}$ belong to the same person and $0$ if they
present different persons. $L^{s}$ is the loss function in the case
of same persons, $L^{d}$ is the loss function in the case of different
ones and $Q$ is a constant representing the upper bound of $E$.

Since the energy is monotonically changing for both $L^{s}$ and $L^{d}$,
the optimization of the loss function can be easily achieved using
simple gradient decent algorithm, and the weights $W$ can be learned
using back-propagation algorithm.

The final step of the proposed system involves detecting the identity
of the query person or verifying his/her identity. This step utilizes
this network structure. For the proposed system, the extracted weights
$P$ from the previous section constiture the feature vector used
in the recognition or verification process. This means that a 1D Siamese
structure network will be used for this verification task. The network
structure is depicted in Figure \ref{fig:Typical-Siamese}. However,
$X_{1}$ and $X_{2}$ will be replaced by vectorized $P_{1}$ and
$P_{2}$ extracted from the previous step. All calculations stated
by equations \eqref{eq:Siamese_energy},\eqref{eq:Siamese_loss},\eqref{eq:Siamese_iter_loss},\eqref{eq:Same_loss},\eqref{eq:diff_loss}
and \eqref{eq:siamese_out} will remain identical. However, $W$ will
consist of 1D vectors as opposed to 2D as in the original case.

\section{Results}

\begin{figure}
\begin{centering}
\includegraphics[scale=0.5]{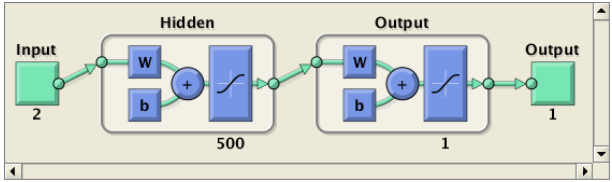}
\par\end{centering}
\caption{Structure of the network used in the experiment. \label{fig:Structure-of-theNetwork}}
\end{figure}

The proposed system has been tested over the Bosphorus \citep{Bosphorus}
database. This dataset consists of 3D and textures for 105 persons
with different facial expressions. For this test only the 3D faces
with neutral expressions are used to test the efficiency of the proposed
neural model. Each person has between 1 to 4 neutral faces with a
total of 299 neutral 3D texture free faces each of them has on average
80,000 3D points. The target mean squared error (MSE) value for the
trained models is below 0.0002. The network structure for this problem
is provided in Figure \ref{fig:Structure-of-theNetwork}. For the
sake of simplicity, only one hidden layer is used in the experiment
with the number of nodes in the hidden layer being 500. The proposed
neural model has been implemented using Neural Network Toolbox of
MATLAB.

\begin{figure}
\begin{centering}
\begin{minipage}[t]{0.45\columnwidth}%
\begin{center}
\includegraphics[scale=0.3]{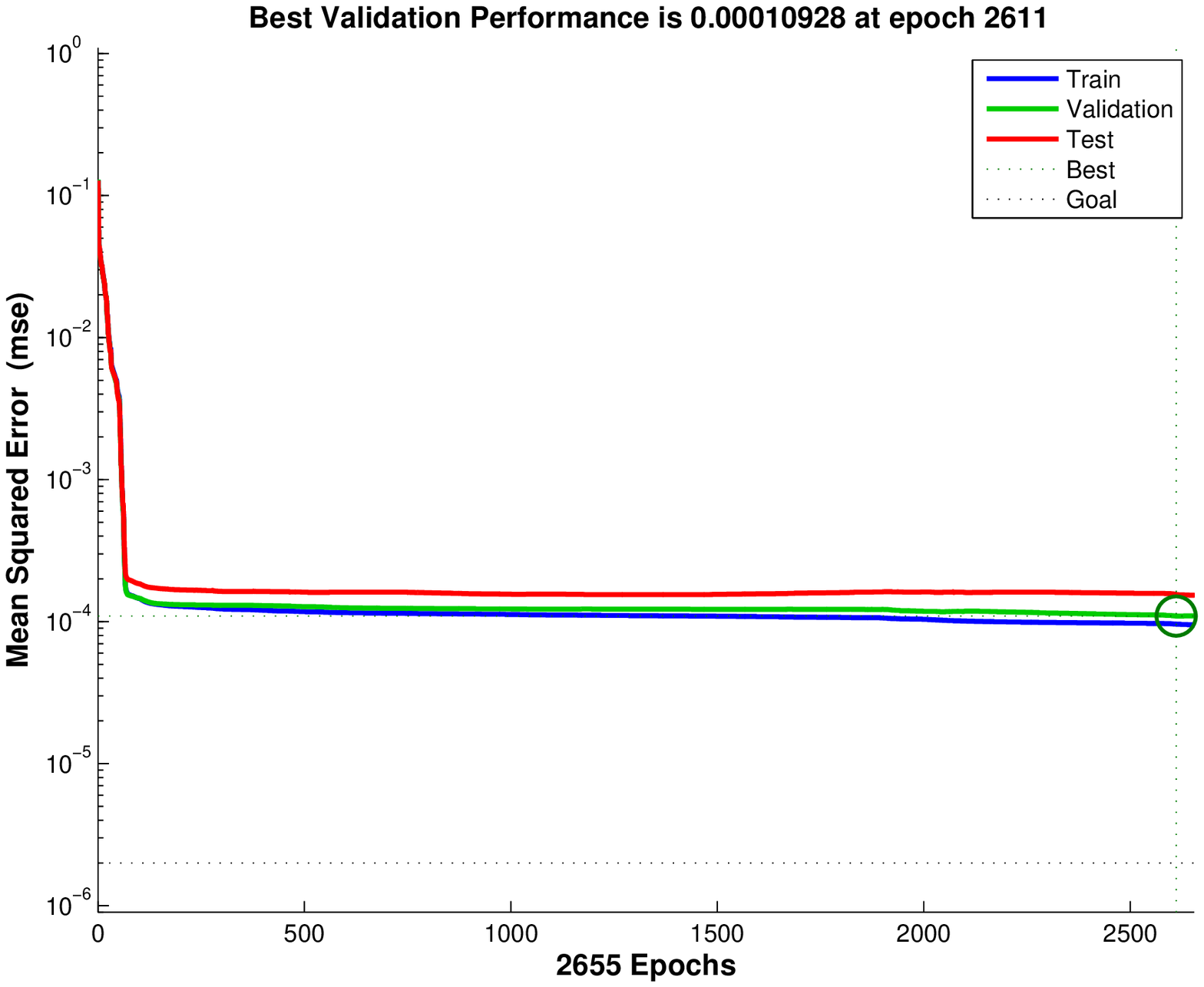} 
\par\end{center}%
\end{minipage}
\par\end{centering}
\begin{centering}
(a)
\par\end{centering}
\begin{centering}
\begin{minipage}[t]{0.45\columnwidth}%
\begin{center}
\includegraphics[scale=0.3]{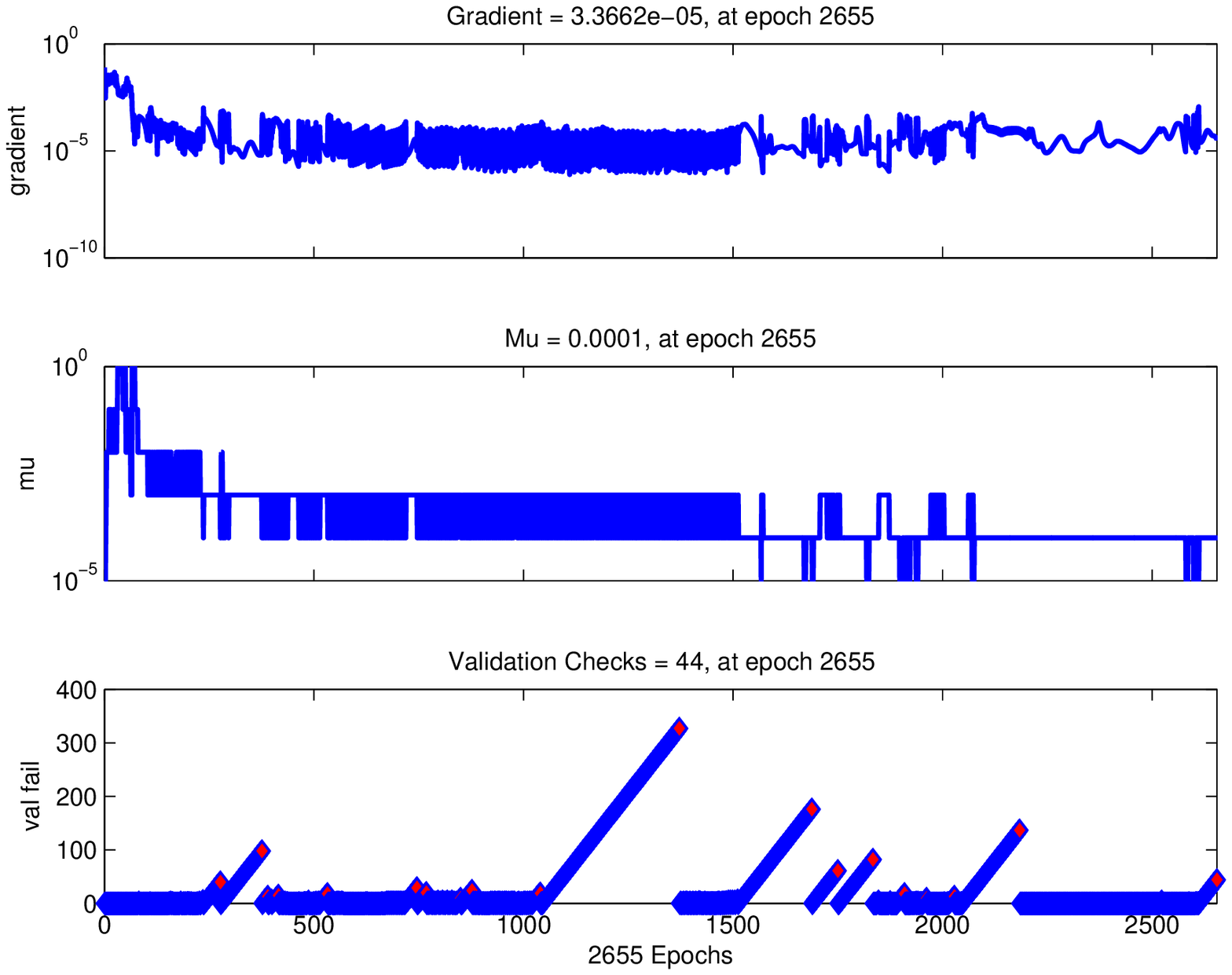} 
\par\end{center}%
\end{minipage}
\par\end{centering}
\begin{centering}
(b)
\par\end{centering}
\caption{Training performance of one of the generated models. \label{fig:Training-Performance-of}}
\end{figure}

As shown in the sample training in Figure \ref{fig:Training-Performance-of},
the training loss improved in the first 100 epochs. Following this,
all upcoming epochs worked as fine tuner for the learning parameters.

\begin{figure}
\begin{centering}
\includegraphics[scale=0.45]{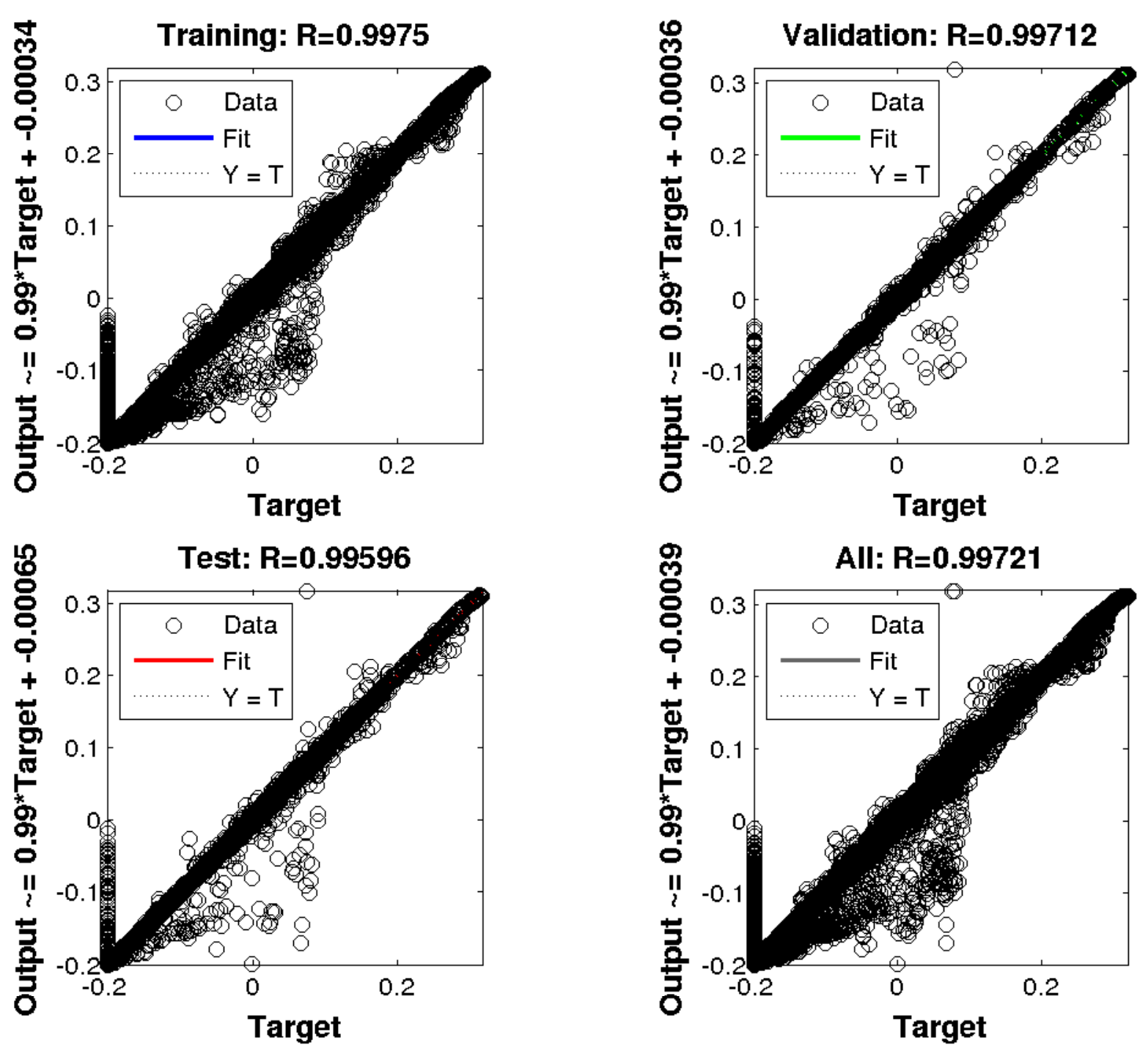}
\par\end{centering}
\caption{Regression result for one of the generated model. \label{fig:Regression-result-for} }
\end{figure}

\begin{figure}
\noindent\begin{minipage}[t]{1\columnwidth}%
\begin{center}
\includegraphics[scale=0.3]{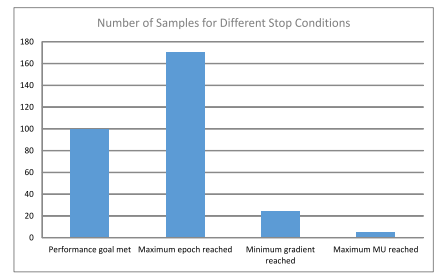}
\par\end{center}%
\end{minipage}

\caption{Number of training models and their stop conditions. \label{fig:Number-of-training}}
\end{figure}

\begin{figure}
\begin{centering}
\noindent\begin{minipage}[t]{1\columnwidth}%
\begin{center}
\includegraphics[scale=0.3]{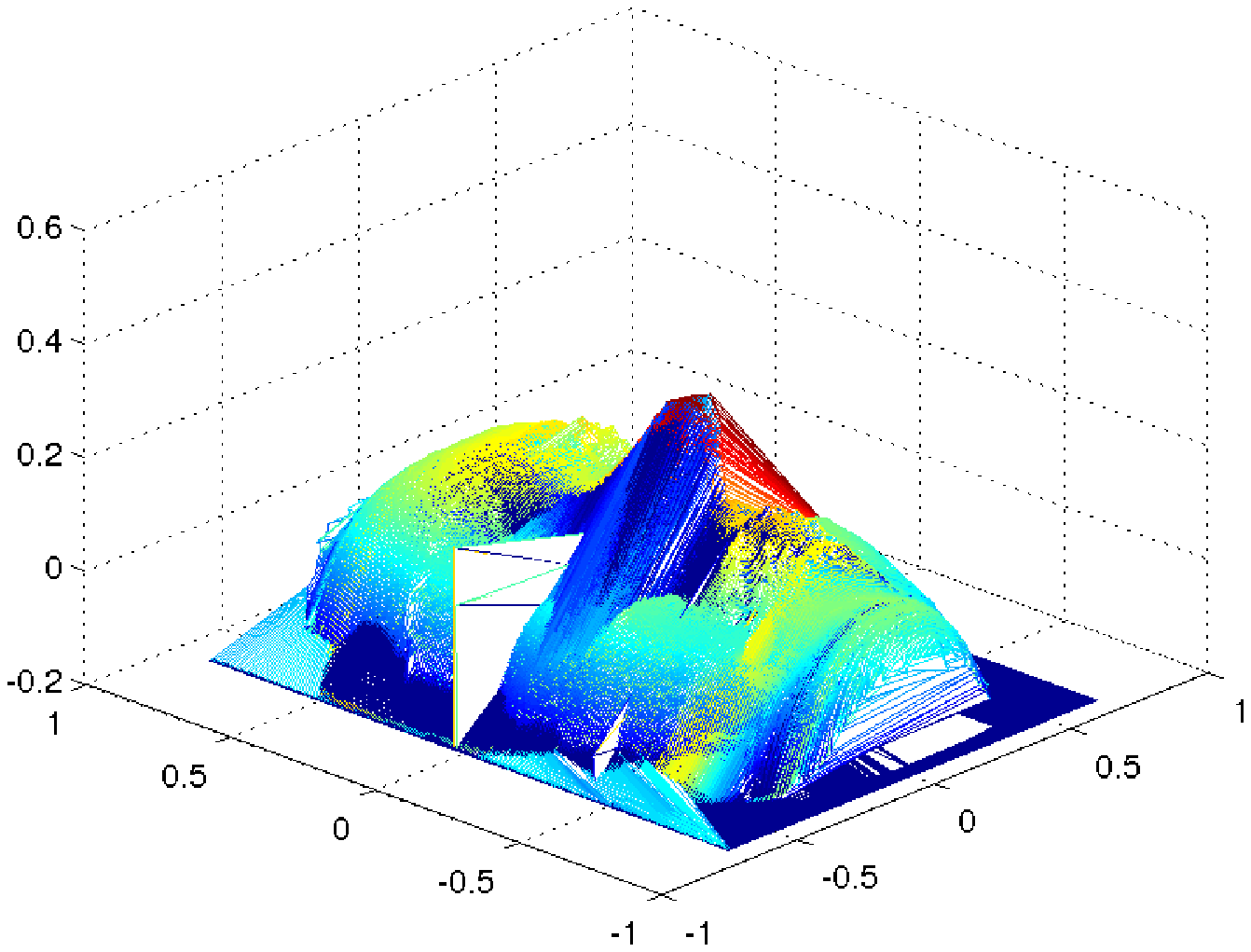}
\par\end{center}%
\end{minipage}
\par\end{centering}
\begin{centering}
(a)
\par\end{centering}
\medskip{}

\noindent\begin{minipage}[t]{1\columnwidth}%
\begin{center}
\includegraphics[scale=0.3]{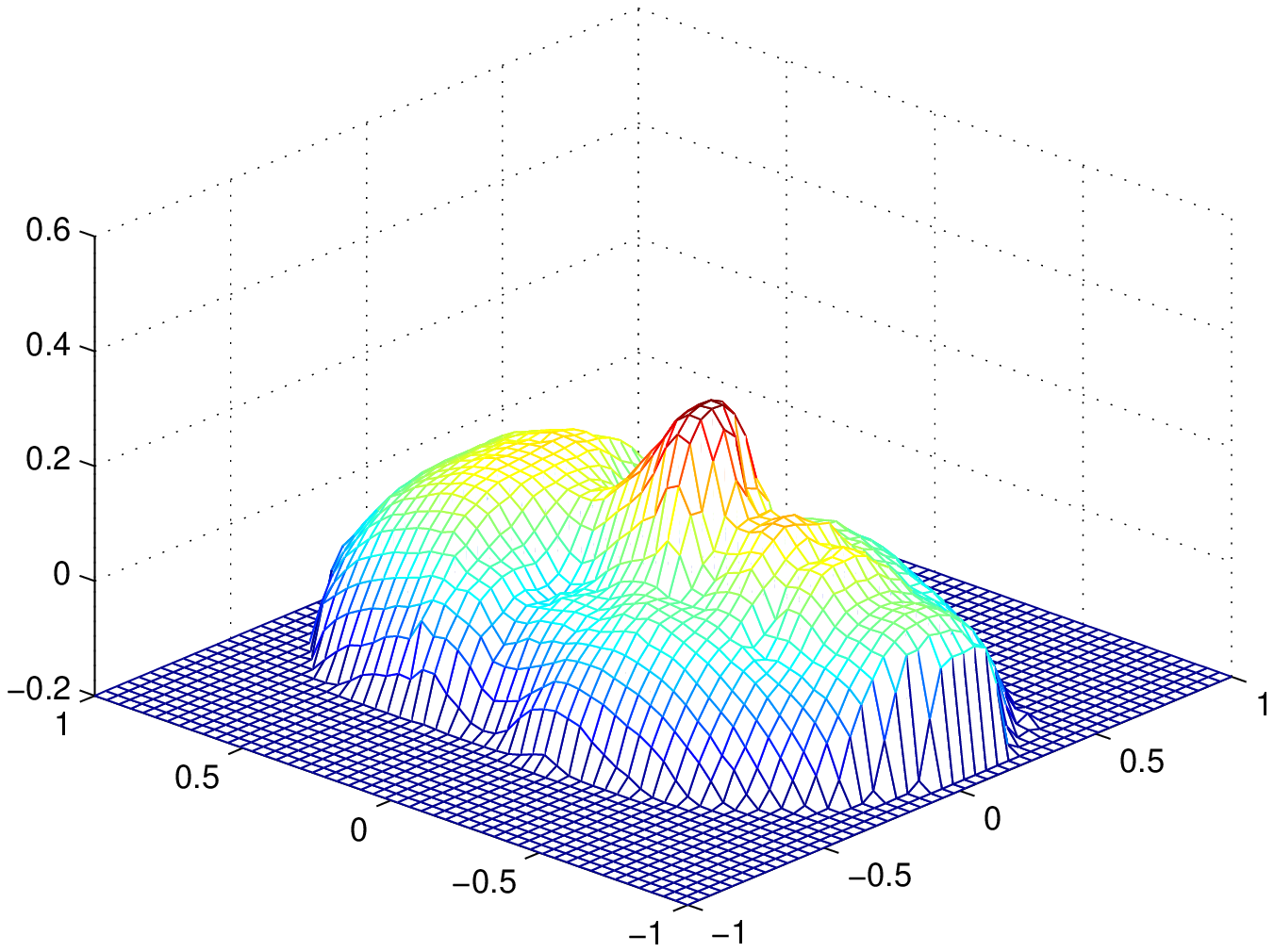}
\par\end{center}%
\end{minipage}
\begin{centering}
(b)
\par\end{centering}
\caption{Sample of original depth points cloud (a) and points cloud generated
by regression model (b). \label{fig:Sample-of-original-generated-depth} }
\end{figure}

A sample of the resulting regression model accuracy is shown in Figure
\ref{fig:Regression-result-for}. It can easily be observed that the
accuracy of the generated model is more than 99\% for presenting target
points cloud. Out of all trained models (299 models), 100 models achieved
the required training MSE of 0.0002, and 170 models terminated when
the maximum number of epochs were reached with an average MSE of 0.00031,
which is still considered to be very low error. Also as shown in Figure
\ref{fig:Number-of-training}, only few models (29 models) terminated
for achieving maximum gradient or maximum $\mu$ value for equation
\eqref{eq:LM_update}. Figure \ref{fig:Sample-of-original-generated-depth}
shows a sample of the generated points cloud compared to the original
one. As also seen in this figure, the original depth points cloud
contains a significantly large noise due to camera and environment.
However, the generated points cloud is smoother and provides better
view of 3D faces.

\begin{figure}
\begin{centering}
\noindent\begin{minipage}[t]{1\columnwidth}%
\begin{center}
\includegraphics[scale=0.2]{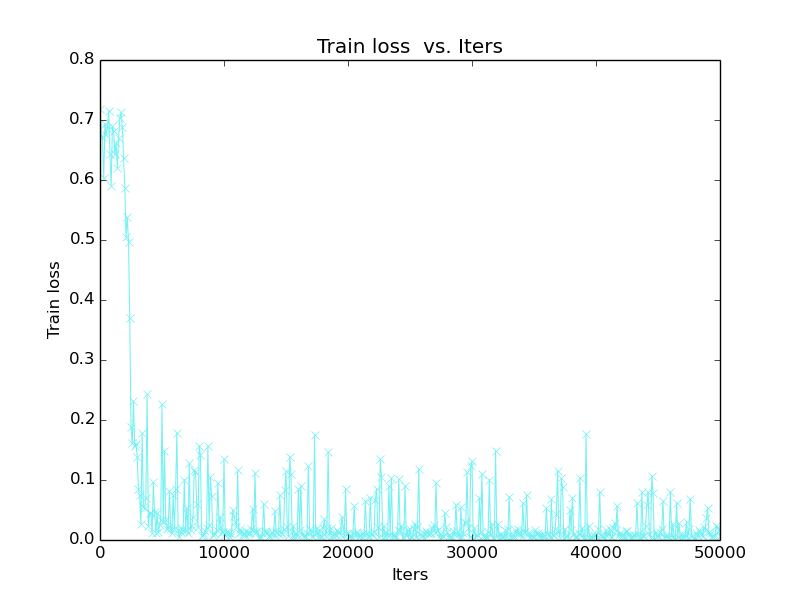}
\par\end{center}
\begin{center}
(a)
\par\end{center}%
\end{minipage}\ %
\noindent\begin{minipage}[t]{1\columnwidth}%
\begin{center}
\includegraphics[scale=0.2]{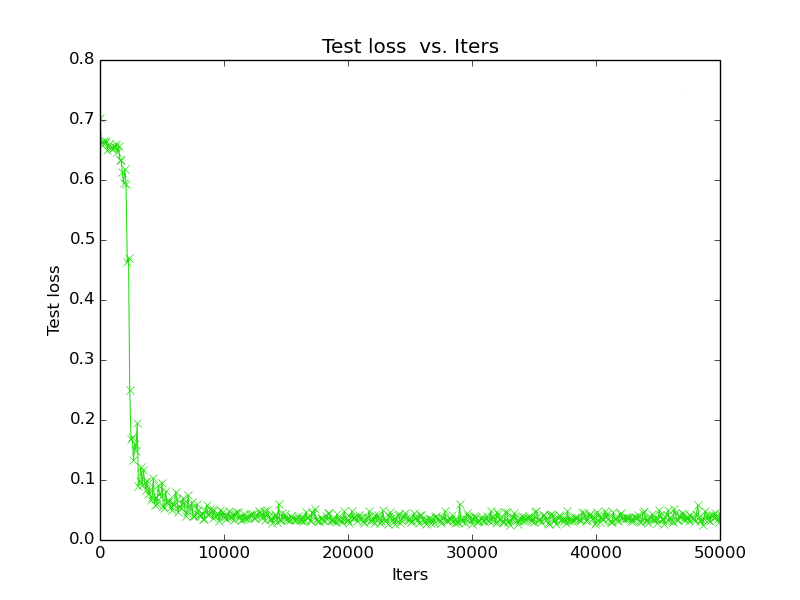}
\par\end{center}
\begin{center}
(b)
\par\end{center}%
\end{minipage}
\par\end{centering}
\medskip{}

\noindent\begin{minipage}[t]{1\columnwidth}%
\begin{center}
\includegraphics[scale=0.2]{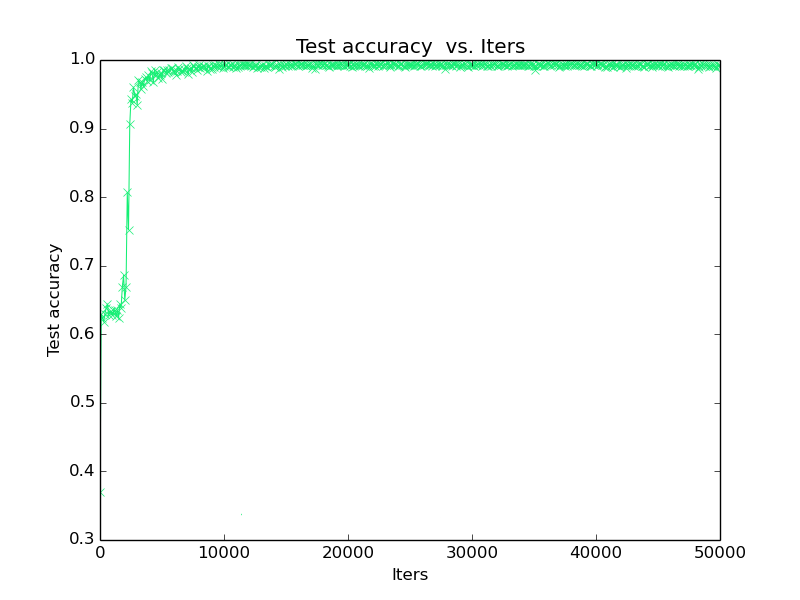}
\par\end{center}
\begin{center}
(c)
\par\end{center}%
\end{minipage}

\caption{(a) Training loss of the Siamese Network over 50000 iterations and
(b) testing loss of the Siamese Network and (c) testing accuracy of
the same network for verification. \label{fig:Siamese_loss_accuracy} }
\end{figure}

\begin{figure}[t]
\begin{centering}
\includegraphics[angle=90,scale=0.12]{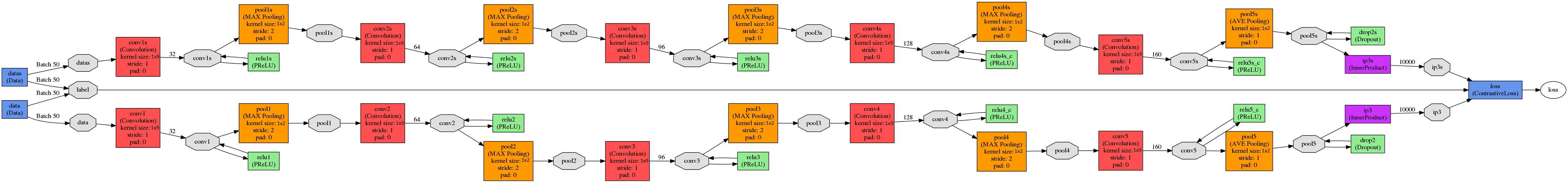}
\par\end{centering}
\caption{Proposed Siamese network structure. \label{fig:Proposed-structure-of-Siamese-3D}}
\end{figure}

\begin{figure}
\begin{centering}
\noindent\begin{minipage}[t]{1\columnwidth}%
\begin{center}
\includegraphics[scale=0.3]{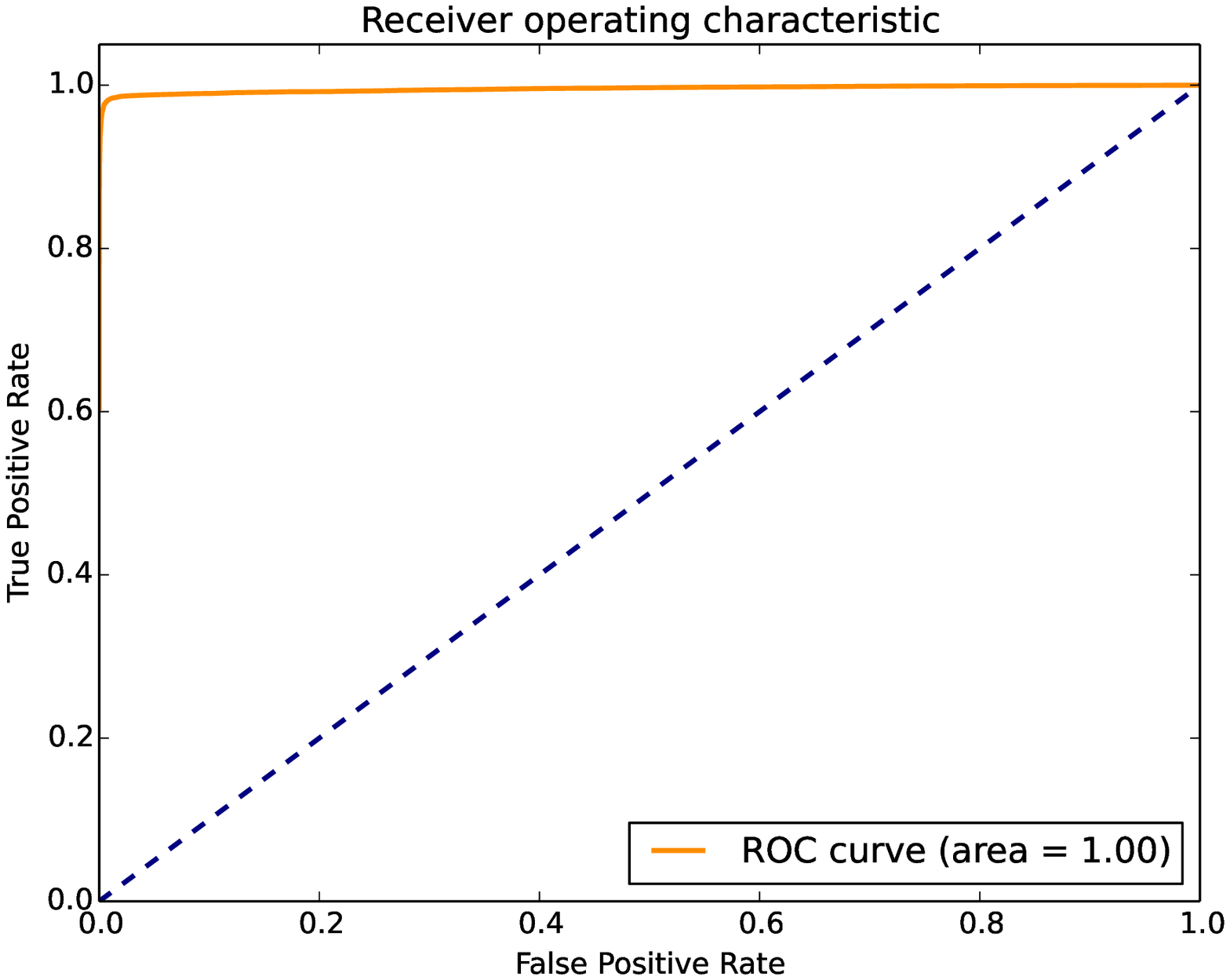}
\par\end{center}%
\end{minipage}
\par\end{centering}
\begin{centering}
(a)
\par\end{centering}
\medskip{}

\noindent\begin{minipage}[t]{1\columnwidth}%
\begin{center}
\includegraphics[scale=0.3]{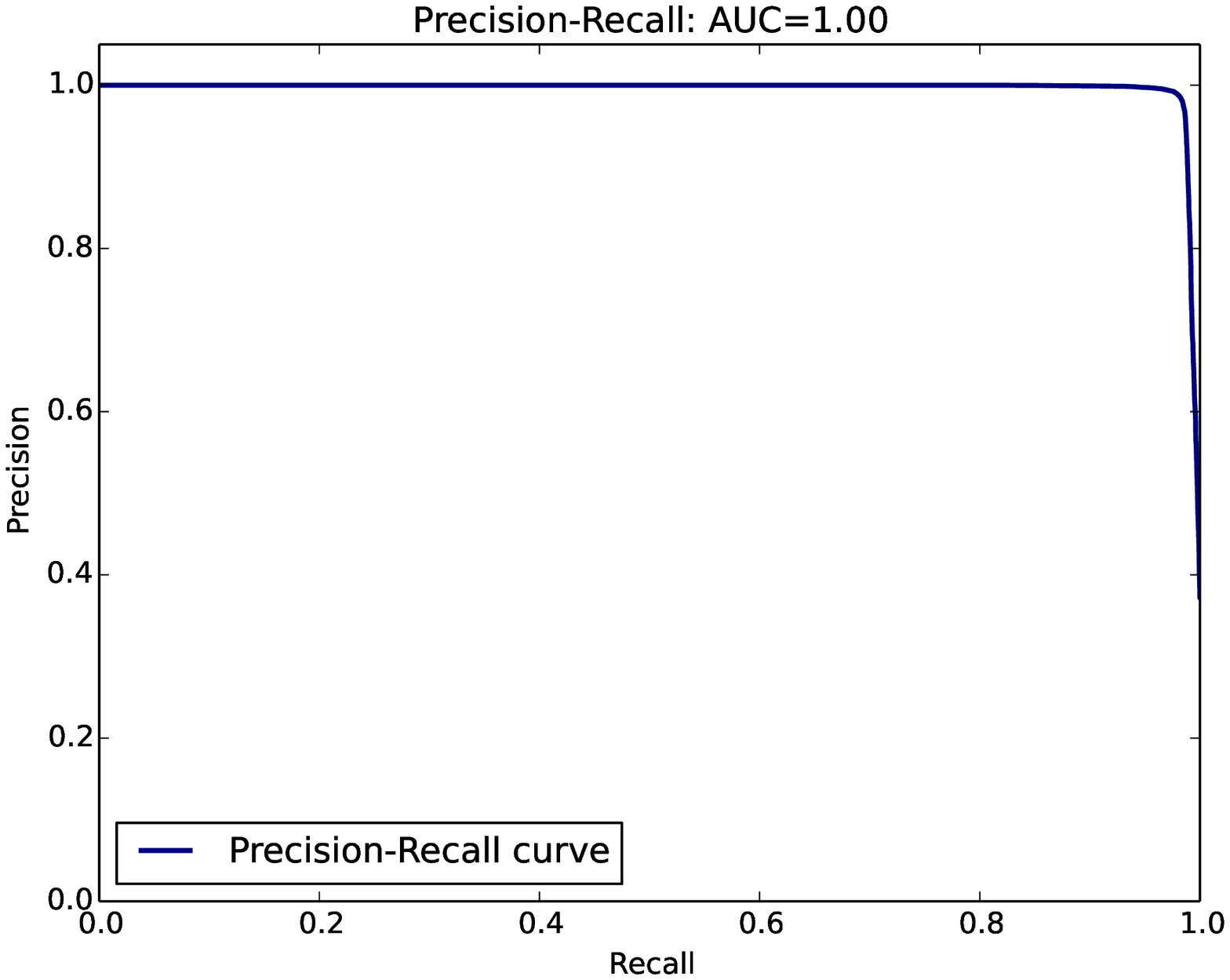}
\par\end{center}%
\end{minipage}
\begin{centering}
(b)
\par\end{centering}
\caption{(a) ROC curve of the Siamese network over testing data and (b) Precision-Recall
curve for the same network on the same data. \label{fig:ROC_Precision-Recall} }
\end{figure}

A Siamese network followed by a verification system has been implemented
using Caffe library with Python wrapper. The structure of the this
network is shown in Figure \ref{fig:Proposed-structure-of-Siamese-3D}.
The 299 neural models which are generated for the 105 persons have
been used for training and testing. Randomly selected pairs from the
generated models have been selected as a positive (pairs for the same
person) and negative (pairs for different persons) training data for
the Siamese network. Since the number of positive samples are so limited,
the data augmentation technique proposed in section 4 has been used
to generate additional pairs for the training and testing. Using this
data augmenting technique, the number of generated samples for same
person class (positive pairs) is 50,000 samples and for different
person class (negative pairs) is 70,000 samples. In the training phase
only 50\% of the generated pairs (50\% of the positive pairs and 50\%
of the negative ones) are used and the remaining 50\% of the data
are used for testing. As it can be seen from Figure \ref{fig:Siamese_loss_accuracy},
the loss function of the trained network did not over-fit, and the
network is efficiently trained. The accuracy of the trained network
over training pairs achieved 100\% and 100\% for testing pairs. The
receiver operation characteristic (ROC) and Precision-Recall curves
for the trained network over the testing pairs are also shown in Figure
\ref{fig:ROC_Precision-Recall}. As also provided in Table \ref{tab:Comparison-of-recognition},
the achieved verification performance is comparable to the state-of-the-art
results published on the same dataset on the neutral expression faces.
These techniques include ICP, Average Regional Models (ARMs), Histogram
of Gradient (HoG), Histogram of Shape index (HoS), Histogram of Gradient
Shape index (HoGS) and the fusion of these histograms (HoG+HoS+HoGS).
Once again, these generated neural models for these faces can regenerate
the 3D point clouds again to be utilized with any of these mentioned
recognition techniques, which give the proposed model more flexibility
if the user want to use or develop different algorithm.

\begin{table}
\begin{centering}
\begin{tabular}{|c|c|}
\hline 
{\small{}Method} & {\small{}Rate (\%)}\tabularnewline
\hline 
\hline 
{\small{}ICP-based holistic approach \citep{Bosph_results}} & {\small{}71.39}\tabularnewline
\hline 
{\small{}Average Regional Models (ARMs) \citep{Bosph_results}} & {\small{}98.87}\tabularnewline
\hline 
{\small{}HoG+HoS+HoGS \citep{Bosph_best}} & {\small{}99.98}\tabularnewline
\hline 
{\small{}proposed Siamese Network} & {\small{}100}\tabularnewline
\hline 
\end{tabular}
\par\end{centering}
\caption{Comparison of recognition rate using different techniques over 3D
neutral expression faces of Bosporus dataset. \label{tab:Comparison-of-recognition}}
\end{table}

\section{Conclusions}

\label{sec:copyright}

In this work, a neural generative modeling technique for texture free
3D faces has been proposed. The neural models have been used for presentation
and regeneration of the 3D faces. The proposed models have been proven
to be accurate representations of the original 3D point clouds with
additional benefits such as reduced 3D cloud storage size, and their
ability to accommodate interpolation and 3D super-resolution. One
other advantage of these models is the flexibility of training from
different number of points in the cloud, which makes them suitable
for heterogeneous cameras. The weights of the trained models have
been used as a raw data with Siamese CNN network for a complete neural
3D face recognition and verification system. These weights have advantage
over the regular 3D clouds for data augmentation. Furthermore, they
allow generation of additional models from a given single model, which
makes this technique advantageous for small dataset recognition given
that one of the limitations of using CNN that it required large dataset
for training and validation. The Siamese network has been trained
over the generated pairs (positive and negative) from the trained
models. The results obtained from the trained Siamese network outperformed
all reported results over the Bosporus dataset for the neutral 3D
faces.

\section*{Acknowledgement}

This work is partially supported by an education research grant from
Amazon Web Services (AWS) as AWS credits.

\bibliographystyle{apalike}
\bibliography{refrences}

\begin{thebibliography}{}

\bibitem[Achermann and Bunke, 2000]{2-3D}
Achermann, B. and Bunke, H. (2000).
\newblock Classifying range images of human faces with hausdorff distance.
\newblock In {\em 2000. Proceedings. 15th International Conference on Pattern
  Recognition,}, volume~2, pages 809--813 vol.2.

\bibitem[Achermann et~al., 1997]{3-3D}
Achermann, B., Jiang, X., and Bunke, H. (1997).
\newblock Face recognition using range images.
\newblock In {\em Proceedings., International Conference on Virtual Systems and
  MultiMedia, 1997. VSMM '97.}, pages 129--136.

\bibitem[Alexandre, 2016]{3D-object-CNNraey}
Alexandre, L. (2016).
\newblock 3d object recognition using convolutional neural networks with
  transfer learning between input channels.
\newblock In Menegatti, E., Michael, N., Berns, K., and Yamaguchi, H., editors,
  {\em Intelligent Autonomous Systems 13}, volume 302 of {\em Advances in
  Intelligent Systems and Computing}, pages 889--898. Springer International
  Publishing.

\bibitem[Alyuz et~al., 2008]{Bosph_results}
Alyuz, N., Gokberk, B., and Akarun, L. (2008).
\newblock A 3d face recognition system for expression and occlusion invariance.
\newblock In {\em 2008 IEEE Second International Conference on Biometrics:
  Theory, Applications and Systems}, pages 1--7.

\bibitem[Besl and McKay, 1992]{ICP}
Besl, P.~J. and McKay, N.~D. (1992).
\newblock A method for registration of 3-d shapes.
\newblock {\em IEEE Trans. Pattern Anal. Mach. Intell.}, 14(2):239--256.

\bibitem[Bowyer et~al., 2006]{3D-survey}
Bowyer, K.~W., Chang, K., and Flynn, P. (2006).
\newblock A survey of approaches and challenges in 3d and multi-modal 3d + 2d
  face recognition.
\newblock {\em Comput. Vis. Image Underst.}, 101(1):1--15.

\bibitem[Bronstein et~al., 2003]{6-3D}
Bronstein, A.~M., Bronstein, M.~M., and Kimmel, R. (2003).
\newblock Expression-invariant 3d face recognition.
\newblock In Kittler, J. and Nixon, M.~S., editors, {\em AVBPA}, volume 2688 of
  {\em Lecture Notes in Computer Science}, pages 62--69. Springer.

\bibitem[Cartoux et~al., 1989]{7-3D}
Cartoux, J.~Y., LaPreste, J.~T., and Richetin, M. (1989).
\newblock Face authentication or recognition by profile extraction from range
  images.
\newblock In {\em Proceedings of the Workshop on Interpretation of 3D Scenes},
  pages 194--199.

\bibitem[Chopra et~al., 2005]{siamese}
Chopra, S., Hadsell, R., and LeCun, Y. (2005).
\newblock Learning a similarity metric discriminatively, with application to
  face verification.
\newblock In {\em CVPR 2005. IEEE Computer Society Conference on Computer
  Vision and Pattern Recognition, 2005.}, volume~1, pages 539--546 vol. 1.

\bibitem[Cretu et~al., 2006]{neural-model}
Cretu, A.-M., Petriu, E., and Patry, G. (2006).
\newblock Neural-network-based models of 3-d objects for virtualized reality: a
  comparative study.
\newblock {\em IEEE Transactions on Instrumentation and Measurement,},
  55(1):99--111.

\bibitem[Daoudi et~al., 2013]{3D-faces-book}
Daoudi, M., Srivastava, A., and Veltkamp, R. (2013).
\newblock {\em 3D Face Modeling, Analysis and Recognition}.
\newblock Wiley Publishing, 1st edition.

\bibitem[Gordon, 1992]{9-3D}
Gordon, G. (1992).
\newblock Face recognition based on depth and curvature features.
\newblock In {\em Proceedings CVPR '92., 1992 IEEE Computer Society Conference
  on Computer Vision and Pattern Recognition, 1992.}, pages 808--810.

\bibitem[Hagan and Menhaj, 1994]{LM-Back}
Hagan, M. and Menhaj, M. (1994).
\newblock Training feedforward networks with the marquardt algorithm.
\newblock {\em IEEE Transactions on Neural Networks,}, 5(6):989--993.

\bibitem[Hesher et~al., 2003]{10-3D}
Hesher, C., Srivastava, A., and Erlebacher, G. (2003).
\newblock A novel technique for face recognition using range imaging.
\newblock In {\em Proceedings. Seventh International Symposium on Signal
  Processing and Its Applications, 2003.}, volume~2, pages 201--204 vol.2.

\bibitem[Kim et~al., 2017]{Deep3DID}
Kim, D., Hernandez, M., Choi, J., and Medioni, G. (2017).
\newblock Deep 3d face identification.
\newblock In {\em 2017 IEEE International Joint Conference on Biometrics
  (IJCB)}, pages 133--142.

\bibitem[Lee and Milios, 1990]{12-3D}
Lee, J. and Milios, E. (1990).
\newblock Matching range images of human faces.
\newblock In {\em Proceedings, Third International Conference on Computer
  Vision, 1990.}, pages 722--726.

\bibitem[Lee et~al., 2003]{13-3D}
Lee, Y., Park, K., Shim, J., and Yi, T. (2003).
\newblock 3d face recognition using statistical multiple features for the local
  depth information.
\newblock In {\em 16th International Conference on Vision Interface}.

\bibitem[Li et~al., 2011]{Bosph_best}
Li, H., Huang, D., Lemaire, P., Morvan, J.~M., and Chen, L. (2011).
\newblock Expression robust 3d face recognition via mesh-based histograms of
  multiple order surface differential quantities.
\newblock In {\em 2011 18th IEEE International Conference on Image Processing},
  pages 3053--3056.

\bibitem[Maes et~al., 2010]{mesh-SIFT}
Maes, C., Fabry, T., Keustermans, J., Smeets, D., Suetens, P., and
  Vandermeulen, D. (2010).
\newblock Feature detection on 3d face surfaces for pose normalisation and
  recognition.
\newblock In {\em 2010 Fourth IEEE International Conference on Biometrics:
  Theory, Applications and Systems (BTAS)}, pages 1--6.

\bibitem[Medioni and Waupotitsch, 2003]{14-3D}
Medioni, G. and Waupotitsch, R. (2003).
\newblock Face recognition and modeling in 3d.
\newblock In {\em IEEE International Workshop on Analysis and Modeling of Faces
  and Gestures (AMFG 2003)}, pages 232--233.

\bibitem[Min et~al., 2003]{16-3D}
Min, J., Bowyer, K.~W., and Flynn, P. (2003).
\newblock Using multiple gallery and probe images per person to improve
  performance of face recognition.
\newblock Technical report, Notre Dame Computer Science and Engineering
  Technical Report.

\bibitem[Min et~al., 2012]{realtime-3D-face}
Min, R., Choi, J., Medioni, G., and Dugelay, J. (2012).
\newblock Real-time 3d face identification from a depth camera.
\newblock In {\em 2012 21st International Conference on Pattern Recognition
  (ICPR),}, pages 1739--1742.

\bibitem[Moreno et~al., 2003]{17-3D}
Moreno, A.~B., Ángel Sánchez, Vélez, J.~F., and Díaz, F.~J. (2003).
\newblock Face recognition using 3d surface-extracted descriptors.
\newblock In {\em In Irish Machine Vision and Image Processing Conference
  (IMVIP 2003), Sepetember}.

\bibitem[Nagamine et~al., 1992]{18-3D}
Nagamine, T., Uemura, T., and Masuda, I. (1992).
\newblock 3d facial image analysis for human identification.
\newblock In {\em Proceedings., 11th IAPR International Conference on Pattern
  Recognition, 1992. Vol.I. Conference A: Computer Vision and Applications,},
  pages 324--327.

\bibitem[Nair and Hinton, 2009]{deep-belief-3D}
Nair, V. and Hinton, G.~E. (2009).
\newblock 3d object recognition with deep belief nets.
\newblock In Bengio, Y., Schuurmans, D., Lafferty, J., Williams, C., and
  Culotta, A., editors, {\em Advances in Neural Information Processing Systems
  22}, pages 1339--1347. Curran Associates, Inc.

\bibitem[{Richard Socher and Brody Huval and Bharath Bhat and Christopher D.
  Manning and Andrew Y. Ng}, 2012]{3D-CRNN}
{Richard Socher and Brody Huval and Bharath Bhat and Christopher D. Manning and
  Andrew Y. Ng} (2012).
\newblock {Convolutional-Recursive Deep Learning for 3D Object Classification}.
\newblock In {\em {Advances in Neural Information Processing Systems 25}}.

\bibitem[Rusinkiewicz and Levoy, 2001]{fast-ICP}
Rusinkiewicz, S. and Levoy, M. (2001).
\newblock Efficient variants of the {ICP} algorithm.
\newblock In {\em Third International Conference on 3D Digital Imaging and
  Modeling (3DIM)}.

\bibitem[Savran et~al., 2008]{Bosphorus}
Savran, A., Aly\"{u}z, N., Dibeklio\u{g}lu, H., \c{C}eliktutan, O.,
  G\"{o}kberk, B., Sankur, B., and Akarun, L. (2008).
\newblock Biometrics and identity management.
\newblock chapter Bosphorus Database for 3D Face Analysis, pages 47--56.
  Springer-Verlag, Berlin, Heidelberg.

\bibitem[Scheenstra et~al., 2005]{3D-survey2}
Scheenstra, A., Ruifrok, A., and Veltkamp, R. (2005).
\newblock A survey of 3d face recognition methods.
\newblock In Kanade, T., Jain, A., and Ratha, N., editors, {\em Audio- and
  Video-Based Biometric Person Authentication}, volume 3546 of {\em Lecture
  Notes in Computer Science}, pages 891--899. Springer Berlin Heidelberg.

\bibitem[Tanaka et~al., 1998]{20-3D}
Tanaka, H., Ikeda, M., and Chiaki, H. (1998).
\newblock Curvature-based face surface recognition using spherical correlation.
  principal directions for curved object recognition.
\newblock In {\em Proceedings. Third IEEE International Conference on Automatic
  Face and Gesture Recognition, 1998.}, pages 372--377.

\end{thebibliography}

\end{document}